\documentclass{Interspeech2024}

\usepackage[T1]{fontenc}
\usepackage[utf8]{inputenc}

\interspeechcameraready

\title{On the Problem of Text-To-Speech Model Selection for\\Synthetic Data Generation in Automatic Speech Recognition\vspace{-0.7em}}

\name[affiliation={1,2,3,*}]{Nick}{Rossenbach}
\name[affiliation={2,3}]{Ralf}{Schlüter}
\name[affiliation={1,4}]{Sakriani}{Sakti}

\address{
  $^1$Japan Advanced Institute of Science and Technology, Japan\\
  $^2$RWTH Aachen University, Germany; 
  $^3$AppTek GmbH, Germany\\
  $^4$Nara Institute of Science and Technology, Japan}
\email{rossenbach@cs.rwth-aachen.de, ssakti@is.naist.jp, schlueter@cs.rwth-aachen.de}


\usepackage{tikz}
\usepackage{multirow}
\usepackage{makecell}
\usetikzlibrary{arrows,positioning} 
\usetikzlibrary{shapes,decorations}
\usetikzlibrary{calc}
\usetikzlibrary{fit}
\usetikzlibrary{decorations.pathreplacing}
\usetikzlibrary{backgrounds}

\definecolor{blind_red}{HTML}{D7191C}
\definecolor{blind_orange}{HTML}{FDAE61}
\definecolor{blind_yellow}{HTML}{FFFFBF}
\definecolor{blind_blue}{HTML}{ABD9E9}
\definecolor{blind_blue2}{HTML}{2C7BB6}

\newcommand\blfootnote[1]{%
  \begingroup
  \renewcommand\thefootnote{}\footnote{#1}%
  \addtocounter{footnote}{-1}%
  \endgroup
}

\begin{document}

\maketitle

\begin{abstract}
The rapid development of neural text-to-speech (TTS) systems enabled its usage in other areas of natural language processing such as automatic speech recognition (ASR) or spoken language translation (SLT).
Due to the large number of different TTS architectures and their extensions, selecting which TTS systems to use for synthetic data creation is not an easy task.
We use the comparison of five different TTS decoder architectures in the scope of synthetic data generation to show the impact on CTC-based speech recognition training.
We compare the recognition results to computable metrics like NISQA MOS and intelligibility, finding that there are no clear relations to the ASR performance.
We also observe that for data generation auto-regressive decoding performs better than non-autoregressive decoding, and propose an approach to quantify TTS generalization capabilities.
\end{abstract}

\section{Introduction}

The usage of synthetic data from text-to-speech (TTS) systems in the context of automatic speech recognition (ASR) has been explored in many different ways.
\blfootnote{*Work done while being an internship student at JAIST}
This ranges from simply generating new data \cite{Laptev-2020-YouDoNotNeedMore,9688255} to jointly training ASR and TTS systems\nobreakspace\cite{8683480}.
Prior work improved the integration of the synthetic data into the ASR systems, for example by different weighting of synthetic inputs \cite{Baskar-2021-EatEnhancedASR-TT}, or the usage of internal ASR model statistics to reject synthetic samples not matching the real data\nobreakspace\cite{Hu-2022-SYNTUtilizingIm}.
As most publications aim to introduce new methodologies, in each case they usually use one specific TTS architecture and one specific ASR architecture.
The number of fundamentally different ASR architectures is rather limited and roughly categorized into: Hybrid deep neural network hidden Markov model (DNN-HMM) \cite{bourlard2012connectionist}, connectionist-temporal-classification (CTC) \cite{Graves06connectionisttemporal}, Transducer \cite{DBLP:journals/corr/abs-1211-3711} and Attention-Encoder-Decoder (AED) \cite{chan2016listen}. For TTS there is a vast amount of different architectures and models, especially when including different approaches of multi-speaker modeling\nobreakspace\cite{Tan2021ASO}.
From the perspective of working on ASR this makes it very difficult to choose which TTS system to use for synthetic data generation.
In addition, human evaluations are not suited to evaluate data for training purposes, as conducting the evaluation usually takes more time than running the training itself. 
Then, evaluation for naturalness and human subjective evaluations alone can not be expected to be a sufficient criterion for utility estimation of synthetic data.
Instead, aspects like speaker diversity, temporal diversity or expressiveness in general are assumed to be relevant to yield good results by integrating synthetic data into the ASR process\nobreakspace\cite{minixhofer23_interspeech,10389782}. Unfortunately, those aspects are difficult to quantify.
Furthermore, LJSpeech \cite{ljspeech} is often the only common dataset to compare new TTS architectures, which can not be expected to give representative results for the multi-speaker case\nobreakspace\cite{wu22f_interspeech}.
This is especially relevant as one might not have training data suited for TTS training at hand for a particular domain or language, but instead has to use the ASR data.

Motivated by those problems, we present an experiment pipeline to systematically analyze the relation of TTS generated data to ASR training.
For this we compare different decoder architectures, which are based on: Non-autoregressive (NAR) Transformer similar to FastSpeech-2 \cite{Ren-2020-FastSpeech2Fasta}, autoregressive (AR) long-short term memory (LSTM) similar to Non-attentive Tacotron \cite{Shen-2020-Non-AttentiveTacotr}, Glow-TTS \cite{kim_glow-tts_2020} as flow-based decoder and Grad-TTS \cite{popov_grad-tts_2021} as diffusion based decoder.
We keep the rest of the TTS systems as simple as possible, using a Transformer encoder and convolution based duration prediction network, as used in \cite{Ren-2020-FastSpeech2Fasta,kim_glow-tts_2020,li_neural_2019}.
Given the large amount of possible TTS architectures, we choose these systems for multiple reasons:
First, they have a widespread recognition in the scientific community. Then, they already share similar encoder and up-sampling concepts, sometimes even using the same code, which makes comparison easier. Finally, they do not require training with an integrated vocoder model (as e.g. in VITS) which would make a direct comparison more difficult.  
While comparisons of different TTS architectures with respect to naturalness and expressiveness exist \cite{zhang23o_interspeech}, this is to our knowledge the first work covering a broader range of different TTS output characteristics and quality levels targeted for ASR training.
Also, we perform a direct comparison of NAR vs AR decoding for data generation with a consistent sequence mapping method.
Lastly, we present a way of measuring generalization of the TTS systems with respect to ASR training data generation by using different conditions for synthesizing the training data. The Sisyphus \cite{DBLP:conf/emnlp/PeterBN18} based setup and additional materials are available on Github\footnote{\scriptsize\url{https://github.com/rwth-i6/i6_experiments/tree/main/users/rossenbach/experiments/jaist_project}}.
\section{Text-to-speech systems}

\subsection{General architecture}

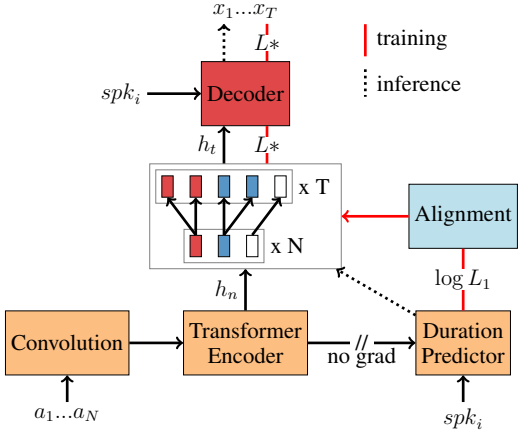
\begin{figure}
\vspace{-2.0em}
	\begin{center}
	\resizebox{2.7in}{!}{
	\begin{tikzpicture}[
	    auto,
	    onlytext/.style={align=center, font=\large},
        halfbox/.style={rectangle, minimum size=1.20cm, draw, font=\large},
	    box/.style={rectangle, minimum size=1.20cm, draw, font=\large},
	    halforangebox/.style={halfbox, fill=blind_orange!80},
	    orangebox/.style={box, fill=blind_orange!80},
	    normalline/.style={->, line width=0.5mm},
	    state/.style={rectangle, draw, minimum height=0.4cm, minimum width=0.2cm},
	]
    \node[halforangebox, align=center] (prenet) at (0, 0) {Convolution};
    \node[onlytext, below=0.5cm of prenet] (inlabel) {$a_1 ... a_N$};
    \node[orangebox, align=center, right=1.0cm of prenet] (encoder) {Transformer\\Encoder};
    \node[orangebox, align=center, right=2.0cm of encoder] (dp) {Duration\\Predictor};
    \node[onlytext, below=0.5cm of dp] (dpspk) {${spk}_i$};
    
    \def \statesep {0.3cm}    
    
    \node[state, above=1.0cm of encoder, fill=blind_blue2!80, xshift=-0.4cm] (estate2) {};
    \node[state, left=\statesep of estate2, fill=blind_red!80] (estate1) {};
    \node[state, right=\statesep of estate2] (estate3) {};
    
    \node [draw=black!50, fit={(estate1) (estate2) (estate3)}] (estatebox) {};
    
    \node [right=0.0cm of estatebox] (estatelabel) {\large x N};
    
    \node[state, above=0.7cm of estate2, fill=blind_blue2!80] (dstate3) {};
    \node[state, left=\statesep of dstate3, fill=blind_red!80] (dstate2) {};
    \node[state, left=\statesep of dstate2, fill=blind_red!80] (dstate1) {};
    \node[state, right=\statesep of dstate3, fill=blind_blue2!80] (dstate4) {};
    \node[state, right=\statesep of dstate4] (dstate5) {};
    
    \node [draw=black!50, fit={(dstate1) (dstate2) (dstate3) (dstate4) (dstate5)}] (dstatebox) {};
    
    \draw [normalline] (estate1.north) -- (dstate1.south); 
    \draw [normalline] (estate1.north) -- (dstate2.south);
    
    \draw [normalline] (estate2.north) -- (dstate3.south); 
    \draw [normalline] (estate2.north) -- (dstate4.south);
    
    \draw [normalline] (estate3.north) -- (dstate5.south);
    
    \node [right=0.0cm of dstatebox] (dstatelabel) {\large x T};
    
    \coordinate (aligncenter) at ($(estate1.north)!.5!(dstate3.south)$);
    
    \node [draw=black!50, fit={(estatebox) (dstatebox) (dstatelabel)}] (upsamplebox) {};
    
    \path let \p1 = (dp), \p2 = (aligncenter) in node at (\x1, \y2) [box, align=center, fill=blind_blue!80] (alignment)  {Alignment};
    
    \node[box,fill=blind_red!80, align=center, above=0.7cm of upsamplebox] (decoder) {Decoder};
    \node[onlytext, left=1cm of decoder] (decspk) {${spk}_i$};
    \node[onlytext, above=0.7cm of decoder] (logmel) {$x_1 ... x_T$};
    
    \draw [normalline] (inlabel) -- (prenet);
    \draw [normalline] (prenet) -- (encoder);
    \draw [normalline] (dpspk) -- (dp);
    \draw [normalline] (encoder) -- ++(dp) node [anchor=center, midway, fill=white] (gradstop) {\large //};
    \node [below=-0.3cm of gradstop] (nogradlabel) {\large no grad};
    \draw [normalline] (encoder) -- (upsamplebox) node [anchor=east, midway] (gradstop) {\large $h_n$};
    \draw [normalline, red] (alignment) -- (upsamplebox);
    \draw [normalline] ([xshift=-0.4cm]upsamplebox.north) -- ([xshift=-0.4cm]decoder.south) node [anchor=east, midway] (targeth) {\large $h_t$};
    \draw [normalline] (decspk) -- (decoder);
    \draw [normalline, dotted] (dp) -- (upsamplebox);
    
    \draw [line width=0.5mm, red] (dp) -- (alignment) node [black, anchor=center, midway, fill=white, inner sep=0] {\large $\log L_1$};
    \draw [line width=0.5mm, red] ([xshift=0.4cm]upsamplebox.north) -- ([xshift=0.4cm]decoder.south) node [black, anchor=center, midway, fill=white, inner sep=0] (mfeloss) {\large $L*$};
    \draw [line width=0.5mm, red] ([xshift=0.4cm]logmel.south) -- ([xshift=0.4cm]decoder.north) node [black, anchor=center, midway, fill=white, inner sep=0] (l1loss) {\large $L*$};
    \draw [normalline, dotted] ([xshift=-0.4cm]decoder.north) -- ([xshift=-0.4cm]logmel.south);
    
		\node[right=1.5cm of decoder, yshift=1.0cm, anchor=west] (trainingline) {\large training};
		\draw [line width=0.5mm, red] ([xshift=-0.1cm]trainingline.north west) -- ([xshift=-0.1cm]trainingline.south west);
		
		\node[below=0.5cm of trainingline.south west, anchor=west] (inferenceline) {\large inference};
		\draw [line width=0.5mm, dotted] ([xshift=-0.1cm]inferenceline.north west) -- ([xshift=-0.1cm]inferenceline.south west);
    
	\end{tikzpicture}
	}
	\end{center}
	\vspace{-0.6cm}
	\caption[TTS Architecture]{The TTS architecture. We change the decoder marked in red for each of the different systems.}
	\label{fig:tts}
	\vspace{-0.4cm}
\end{figure}

In the design of the TTS systems we follow what we found to be most prevalent in recent literature. After the introduction of Transformer blocks with convolution for TTS \cite{li_neural_2019} and the establishing of duration based state up-sampling \cite{Ren-2019-FastSpeechFastRo}, many systems continued using these components. Thus, for each system we use the same layout of a Transformer based-encoder, a convolution based duration predictor and a discrete encoder state up-sampling method. Only the decoder, which converts the up-sampled encoder states $h_1^T$ into the final log-Mel spectrograms $x_1^T$, is changed. Figure \ref{fig:tts} shows the TTS architecture used in this work. We do not use explicit $f_0$ or energy prediction. The overall structure follows exactly Glow-TTS, in which many parts were taken from TransformerTTS \cite{li_neural_2019} or FastSpeech-2. We also used the code of the official Glow-TTS repository as reference for our implementation.
\subsection{Decoder variants}

A large amount of proposed TTS decoder architectures can be categorized into three groups: Direct prediction, flow and diffusion. Direct prediction means that the output of the TTS neural network is a log-Mel spectrogram which is trained to directly match the target $x_1^T$. Common losses are mean absolute error or mean squared error. The generation is a single forward pass through the decoder network $g_{\theta}$, which can be written as:
\begin{equation}
\hat{x}_1^T = g_{\theta}(h_1^T)
\end{equation}
The generation can also be done in an auto-regressive manner:
\begin{equation}
\hat{x}_t = g_{\theta}(h_1^t, \hat{x}_1^{t-1})
\end{equation}
For flow-based TTS systems, the decoder is an invertible neural network that can be applied in both directions from the spectrograms $x_1^T$ into latent space variables $z_1^T$ and vice-versa. During training, the target features are transformed into the latent space and matched against a normal distribution with a mean $\mu_1^T = \mu_{\theta}(h_1^T)$ and unit variance. The target objective is to maximize the likelihood of the latent space samples given the distributions. For inference, the latent space values are sampled from the distribution with a mean function $\mu_{\theta}$, temperature $\tau$, and passed backwards through the flow $f_{\theta}$:
\begin{equation}
\hat{z}_1^T \sim \mathcal{N}(\mu_{\theta}(h_1^T),\tau\mathbf{1}) \quad and \quad \hat{x}_1^T = f^{-1}_{\theta}(\hat{z}_1^T)
\end{equation}
For the diffusion-based TTS system we follow the Grad-TTS architecture. A normal distribution as in the flow-based approach is used to sample a noisy output spectrogram $\hat{X}_I$ with $X = x_1^T$. The noise spectrogram is cleaned in $I$ steps using a differential equation containing the solver network $s_{\theta}$ and pre-defined noise schedule $\beta_i$ to reach the final output $\hat{X}_0$:
\begin{align}
\hat{X}_I & \sim \mathcal{N}(\mu_{\theta}(h_1^T),\tau\mathbf{1})\\
\hat{X}_{i-1} &= \hat{X}_i - \frac{1}{2I}\left(\mu_{\theta}(h_1^T) - \hat{X}_i - s_{\theta}(h_1^T, \hat{X}_i)\right)\beta_i
\end{align}
For each of the decoder principles we implemented one model for which the details will be presented later in Section \ref{sec:trainandmodel}.




\section{Evaluation pipeline and metrics}



%


To evaluate the TTS systems in the context of ASR, we train a CTC-based ASR system purely on data generated by TTS. We generate the same amount of data  as used for the baseline training, and measure the performance of the ASR systems in word error rate (WER) for recognizing real human speech. In the optimal case, the performance of the ASR system trained on the synthetic data would match the performance of the system trained on real data. In order to evaluate the generalization capabilities of TTS towards new conditions, we generate the synthetic data in 3 ways: a) using exactly the same text and speaker assignment as in training, b) using the same text but shuffling the speakers and finally c) using the same amount of text from a different corpus in the same domain. That way we can measure how good the TTS systems are at generalizing to new conditions different from the training. The main evaluation criterion to measure the utility of the TTS-generated data is the WER of the trained ASR systems calculated on the official LibriSpeech \cite{librispeech}  test sets. In addition, we use metrics directly on the synthetic data to evaluate their relation to the final WER. We use NISQA \cite{mittag20_interspeech} as a tool to generate automatic mean-opinion-score (MOS) values. While computer generated MOS values are not a replacement for subjective evaluations, they can be used as tool to determine naturalness as possible indicator for the ASR performance. Subjective evaluation itself can not be useful to evaluate synthetic data in a development process, as doing the evaluation often takes more effort than simply training the ASR system. We also use a larger ASR model trained on the full LibriSpeech corpus to recognize the synthesized cross-validation data to check for completeness and pronunciation correctness. We refer to this metric as synthetic WER (sWER). In the literature this is often called \textit{intelligibility} metric, and sometimes character error rate is used instead of WER , e.g. in \cite{9053512}.

\begin{table}
\caption{TTS model parameter count, training time on Nvidia A40 GPU and inference real-time-factor on AMD Epyc 7H12 CPU using 2 physical cores and including Griffin \& Lim.}
\label{tab:sizes}
\vspace{-0.5cm}
\center{
\begin{tabular}{|c|c|c|c|}
\hline
\multirow{2}{*}{TTS Decoder} & \multirow{2}{*}{\# params} & \multirow{2}{*}{\makecell{training\\time}} & \multirow{2}{*}{\makecell{CPU\\RTF}}\\
&&&\\
\hline
Transformer & 24.4M & 70h & 0.069\\
NAR LSTM & 29.3M & 27h & 0.025\\
AR LSTM & 33.6M & 94h & 0.028\\
Glow-TTS & 57.3M & 53h & 0.036\\
Grad-TTS & 25.5M & 61h & 0.866\\
\hline
\end{tabular}
}
\vspace{-0.5cm}
\end{table}



\section{Experiments}

\subsection{Data}

We use the LibriSpeech train-clean-100 as baseline data to train the TTS and ASR models, and the transcriptions of train-clean-360 as additional text data. This is a common setup that has been used extensively in previous literature \cite{Laptev-2020-YouDoNotNeedMore,9688255,Baskar-2021-EatEnhancedASR-TT}. We sub-sample the exact same amount of utterances, roughly 28k, from train-clean-360 to have the same amount of unseen text as text in the training data. We take 4 utterances of each of the 251 speakers in train-clean-100 out of the TTS training data to have a cross-validation set to report MOS and sWER. We perform no further pre-processing of the data, unlike other work \cite{9688255,minixhofer23_interspeech}. I.e. no excessive silence is removed, and no speakers are discarded. We use the official LibriSpeech lexicon for word-to-phoneme conversion and use Sequitur \cite{Bisani-2008-Joint-sequencemodel} to generate phonemes for words that are not in the lexicon. For ASR, we augment the phoneme set with end-of-word markers as presented in \cite{9413648}.

\subsection{Automatic speech recogition}
As ASR system we use a CTC model with a Conformer-based\nobreakspace\cite{gulati20_interspeech} encoder with 12 blocks, a base dimension of 384 and a convolutional frontend with a sub-sampling factor of 4. For recognition we use the lexicon-constrained Flashlight \cite{pmlr-v162-kahn22a} CTC decoder via the available Torchaudio interface. The search process includes the official LibriSpeech 4-gram LM model. Since it has been shown that synthetic data generation is successful mostly for AED-ASR \cite{9688255}, we chose the context-free phoneme-based CTC-ASR specifically for the reason that it should be influenced less by changing or adding new text.

\subsection{Training and model settings}
\label{sec:trainandmodel}

All models are trained using 80-dimensional log-Mel spectrogram features, for TTS with 12.5ms shift and for ASR with 10ms shift and SpecAugment \cite{park19e_interspeech} applied.
The TTS models are trained with Adam \cite{DBLP:journals/corr/KingmaB14} for 400 epochs. For all models except Grad-TTS we use linear learning rate (LR) scheduling, which we found to perform better than inverse square reduction. We increase the LR from 5e-5 to 5e-4 for the first 100 epochs, and reduce linearly from 5e-4 to 5e-7. For Grad-TTS we use a constant LR of 1e-4. The ASR models are trained using AdamW  with weight-decay 1e-3 for roughly 80 epochs and a maximum learning rate of 7e-4. We generate an alignment by training the Glow-TTS system once using implicit alignment search. In each TTS training the fixed alignment is used for up-sampling and as target for the duration predictor. For vocoding, we convert the log-Mel to linear features using a separate BLSTM-based network and apply the Librosa Griffin \& Lim (G\&L) \cite{DBLP:conf/icassp/GriffinDL84} function with 32 iterations and momentum of 0.99.

\subsubsection{TTS parameter settings}

For the different decoder architectures the number of parameters, training time and inference time differ, as shown in Table \ref{tab:sizes}. Making a fair comparison is difficult, as for each system we would need to do extensive tuning experiments until we reach the optimal performance. Thus, we stick mostly to the sizes presented in previous publications.
Due to the high cost of subjectively evaluating TTS systems, ablation studies on the model sizes are usually not presented in scientific publications.

\textbf{Base model:} For the TTS Transformer encoder we use 6 layers with an internal dimension of 256, preceded by 3 convolution layers with 256 channels and kernel size 5. The duration prediction network uses 2 convolutions with kernel size 3 and 384 channels.

\textbf{Transformer decoder:} For the Transformer-based decoder we simply re-use the same structure as for the encoder Transformer, having 6 layers with an internal base dimension of 256.
The system is trained using L1 loss on the output spectrograms.

\textbf{Autoregressive LSTM decoder:} For the AR-LSTM decoder we are using 2 layers of 1024-dimensional Zoneout-regularized LSTM layers. We use a reduction factor of 2 to predict two feature frames per decoder step. The pre-net for the AR feedback loop and the convolutional NAR post-net are derived from \cite{Shen-2020-Non-AttentiveTacotr}. We apply L1 loss for both the decoder output and the combined prediction with the post-net.

\textbf{Non-autoregressive LSTM decoder:} This decoder is similar to the AR decoder, but we remove the AR feedback loop network, and replace the uni-directional Zoneout LSTM cell by a normal bi-directional LSTM with dropout. Each LSTM direction has 512 dimensions. For Transformer, AR- and NAR-LSTM, a 256-dimensional speaker embedding vector is concatenated to each upsampled encoder state.

\textbf{Glow-TTS decoder:} For a flow-based decoder, including speaker conditioning, we follow exactly the architecture of Glow-TTS \cite{kim_glow-tts_2020}. The decoder consists of 12 invertible coupling blocks with an internal dimension of 256. We merge each 2 consecutive frames for a latent-space dimension of 160.

\textbf{Grad-TTS decoder:} For a diffusion-based model we follow the architecture and settings of Grad-TTS \cite{popov_grad-tts_2021}. As the original architecture is not designed for multi-speaker TTS, modifications were necessary for convergence. We added two additional Transformer layers on top of the 6 encoder layers which are conditioned on the speaker embedding. That way, we have a speaker-conditioned mean prediction network. We also used the same speaker embeddings to condition the u-net, which is used for the differential equation solving.
We found $\tau = 0.7$ to be the optimal noise scale for both Glow-TTS and Grad-TTS.

\section{Results}{

\subsection{Synthetic data, automatic MOS and recognizability}

\begin{table}
\caption{Evaluation of the different TTS decoder architectures. We use 28k sequences from train-clean-360 for the synthetic training data generation and the cross-validation subset to evaluate NISQA MOS and the sWER. The reference values are the training / evaluation on the real audio. For MOS we include the 95\% confidence interval using the bootstrap method from \cite{Confidence_Intervals}.}
\label{tab:compare1}
\vspace{-0.5cm}
\center{
\begin{tabular}{|l|c|c|c|c|}
\hline
\multirow{3}{*}{TTS Decoder} &  \multicolumn{2}{c|}{Syn. data ASR} & \multirow{3}{*}{\makecell{NISQA\\MOS$\uparrow$}} & \multirow{2}{*}{\makecell{sWER}} \\
\cline{2-3}
& \multicolumn{2}{c|}{test [WER \%]$\downarrow$} & & \multirow{2}{*}{[\%]$\downarrow$}\\ 
\cline{2-3}
& clean & other &  &\\ 
\hline
\hline
Transformer & 11.8 & 33.7 & 2.74 $\pm$ 0.05 & 1.6\\
NAR LSTM & 10.2 & 30.5 & 2.87 $\pm$ 0.05 & 1.7 \\
AR LSTM & \phantom{0}9.4 & 28.4 & 3.08 $\pm$ 0.05 & 2.5\\
Glow-TTS & 13.8 & 34.9 & 3.33 $\pm$ 0.04 & 10.8\\
Grad-TTS & 11.5 & 30.0 & 2.47 $\pm$ 0.05 & 15.1\\
\hline
\hline
Reference & \phantom{0}5.1 & 14.9 & 4.06 $\pm$ 0.07 & 1.6\\
\hline
\end{tabular}
}
\vspace{-0.6cm}
\end{table}

%
In a first comparison we evaluated the systems via the synthetic data only ASR training, the NISQA MOS and the sWER.
Table \ref{tab:compare1} shows the results of each of the five decoders compared to the reference values of the real data. For all metrics we used text data unseen during training.
With respect to the final ASR training evaluation, all TTS systems have a reasonable performance and there is no strong outlier.
In general, using the synthetic data for training results in a system with double the error rate when training TTS only on the available ASR data, which is a consistent result to previous literature \cite{Hu-2022-SYNTUtilizingIm, minixhofer23_interspeech}.
\cite{anastassiou2024seedtts} showed that even when using vast amounts of training data to create strong foundation-model sized TTS models, the generated synthetic data seems to be not yet on par with real data.
The AR-LSTM decoder produces the best synthetic training data, which is unexpected as the Glow-TTS and Grad-TTS should produce more expressive and variable audio due to their internal sampling process.
Also, Glow-TTS is the only system that exactly follows the given literature, while the other systems have small adaptations or are simplifications of the original model architecture.
The third and fourth column in Table \ref{tab:compare1} show the NISQA MOS and the sWER on the TTS cross validation set. It is visible that there is no direct relation of the two metrics compared to the final ASR performance.
For example, when comparing Grad-TTS to the Glow-TTS and Transformer decoder, there is a lower MOS and a much higher sWER, but still the final system performance is better. This means both metrics are not reliable to determine which system to pick for data generation.

\subsection{Decoder characteristics}
In Figure \ref{fig:spectrogram} we show a selected sequence synthesized by four different decoders. The Transformer decoder spectrograms are smooth, lacking a lot of detail in the high-frequency area. The AR-LSTM spectrogram in comparison has much more details. The spectrogram from Glow-TTS shows visible noise resulting from the latent space sampling. It also has less distinct boundaries between the phonemes, but rather fuzzy transitions, which might be an explanation for the worse recognizability. The Grad-TTS has less distorted spectrograms than Glow-TTS, but we found that for very short sequences the output is sometimes just a noisy grid pattern. Also the voice lines become very blurred together for lower frequencies, causing a very unnatural sound, which matches with the low automatic MOS score. Thus we assume, that the NISQA MOS and the sWER are very sensitive to certain artifacts that have less impact in on the actual ASR training. One can see that all four TTS decoders produce spectrograms with different characteristics, even with other components being identical. Thus, it is important to not only consider a single TTS architecture when trying to show a general relation between the ASR performance and metrics calculated on the level of the TTS system or the synthetic data.

\begin{figure}
\begin{center}
\includegraphics[scale=0.85]{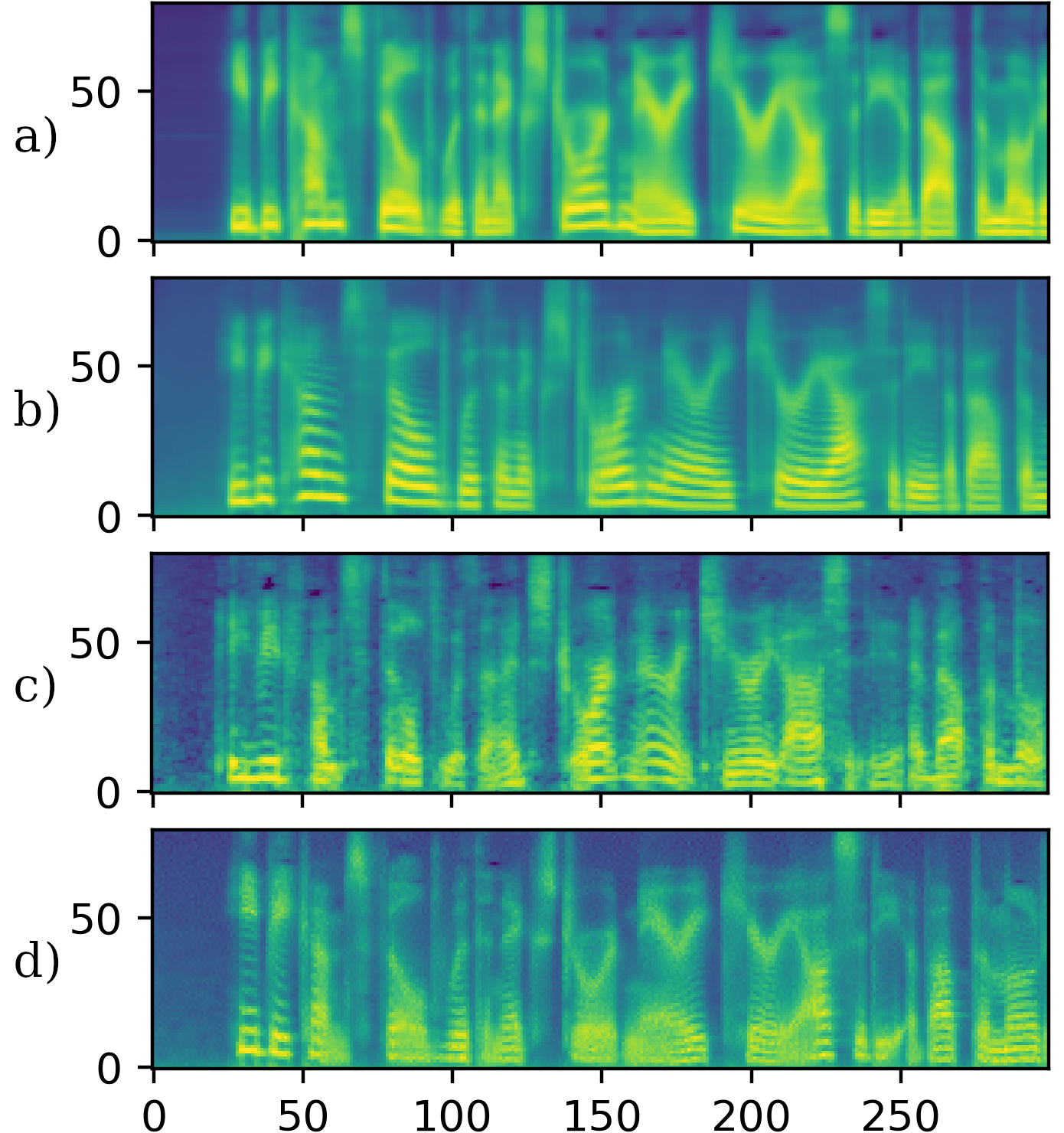}
\caption{Example spectrograms of a selected sequence from the cross validation set. Used TTS decoder from top to bottom: a) Transformer b) AR-LSTM c) Glow-TTS d) Grad-TTS}
\label{fig:spectrogram}
\end{center}
\vspace{-1.0cm}
\end{figure}

%
%

\subsection{Generalization}

\begin{table}
\caption{Comparison of the different TTS decoder architectures regarding overfitting conditions by synthesizing data with \textbf{a)} same text and same speaker as in training \textbf{b)} same text and shuffled speakers \textbf{c)} new text of same amount.}
\label{tab:overfit}
\vspace{-0.5cm}
\center{
\begin{tabular}{|l|c|c|c|c|c|c|}
\hline
\multirow{3}{*}{TTS Decoder} & \multicolumn{6}{c|}{test [WER \%]$\downarrow$} \\
\cline{2-7}
& \multicolumn{3}{c|}{clean} & \multicolumn{3}{c|}{other}\\ 
\cline{2-7}
&a) & b) & c) & a) & b) & c)\\
\hline
\hline
Transformer & 11.1 & 11.8 & 11.8 & 32.5 & 33.5 & 33.7\\
NAR LSTM & \phantom{0}9.7 & 10.1 & 10.2 & 30.0 & 30.2 & 30.5\\
AR LSTM & \phantom{0}8.8 & \phantom{0}9.1 & \phantom{0}9.4 & 26.3 & 27.1 & 28.4\\
Glow-TTS & 12.6 & 13.3 & 13.8 & 31.0 & 32.9 & 34.9\\
Grad-TTS & 11.0 & 11.5 & 11.5 & 28.0 & 30.1 & 30.0\\
\hline
\hline
Reference & \multicolumn{3}{c|}{5.1} & \multicolumn{3}{c|}{14.9}\\
\hline
\end{tabular}
}
\vspace{-0.5cm}
\end{table}

Table \ref{tab:overfit} shows how good the TTS systems generalize with respect to the synthetic data generation task. Recall that condition a) means to exactly reproduce the training data, so with perfect memorization during training it would be possible to produce identically performing data. It is visible that across all models the values for a) and c) are mostly not too far from each other, so the generalization abilities are high. This gives a hint that current models are under-fitting and much larger models might yield better results, even on very constrained datasets such as LibriSpeech 100h. For synthesizing new text, so comparing b) to c), most models show no or only a small change. Only for the AR LSTM and Glow-TTS decoder there is meaningful degradation in the WER, mostly for the "other" test. For the Glow-TTS a possible reason could be that it suffers from pronunciation issues, which get even stronger when synthesizing new text. Testing for the generalization capabilities of TTS this way might not be used only for the comparison of different models like in this work, but allow for easier hyperparameter tuning which is usually not done extensively in TTS literature.

\subsection{Discussion}

When training with synthetic data only, we achieved a WER performance which is less than factor 2 worse compared to training on the real data. In relation, \cite{Hu-2022-SYNTUtilizingIm} reported reaching ratios of around 2 on the LibriSpeech-960 task, and \cite{minixhofer23_interspeech} reported ratios of more than 3. In this work we specifically investigated decoder architectures. In future work this gap can be reduced by extending the results of this work with extensive hyperparameter tuning as well as prosody \cite{minixhofer23_interspeech} and more sophisticated duration modeling \cite{10389782}. We have seen that NISQA MOS and the sWER are not reliable to determine the usefulness of the TTS system for ASR data generation. This means there are more and/or different metrics needed to give accurate estimates about how good the synthetic data can be utilized for ASR training. While in our work synthesizing new data and training the ASR systems was done in less than a day, this becomes much more of a problem when training large models on magnitudes more of data. Thus, we think it is important to find ways to estimate the capabilities of TTS models before starting large-scale data generation.

\section{Conclusion}
In this work we evaluated different TTS decoder architectures w.r.t. their ability to generate synthetic data for ASR training. We compared five different TTS decoder architectures and found that for our setup the AR-LSTM-based system with L1 loss performed best in creating training data for ASR. We were able to create synthetic data that performs less than a factor of 2 worse than real data while training the TTS only on the same data as available for the ASR training. We observed that the systems' outputs differ from each other in many aspects such as naturalness or pronunciation accuracy. Still, the metrics we used to capture them showed a surprisingly low correlation to the performance of the ASR systems trained on such outputs. This indicates that better metrics have to be found in order to do a meaningful evaluation of TTS systems before using it to create synthetic training data. We also introduced a way to measure the generalization capabilities of TTS systems for data generation, which can be used to optimize TTS towards being able to generate data that reaches parity with real data.

\section{Acknowledgements}
\vspace{-0.5em}
{
\footnotesize
This work was partially supported by NeuroSys, which as part of the initiative “Clusters4Future” is funded by the Federal Ministry of Education and Research BMBF (03ZU1106DA), and by the project RESCALE within the program \textit{AI Lighthouse Projects for the Environment, Climate, Nature and Resources} funded by the Federal Ministry for the Environment, Nature Conservation, Nuclear Safety and Consumer Protection (BMUV), funding ID: 67KI32006A. Part of this work is supported by JSPS KAKENHI Grant Numbers JP21H05054 and JP23K21681. The authors thank Lukas Rilling, Peter Vieting and Benedikt Hilmes for their additional input and feedback.
}
\vspace{-0.5em}

\bibliographystyle{IEEEtran}

\let\OLDthebibliography\thebibliography
\renewcommand\thebibliography[1]{
  \OLDthebibliography{#1}
  \setlength{\parskip}{0.1pt}
  \setlength{\itemsep}{0.25pt}
}

\bibliography{mybib}

\end{document}